\documentclass[conference,10pt,letterpaper]{IEEEtran}
\usepackage{times}
\usepackage[numbers,sort&compress]{natbib}
\usepackage{multicol}
\usepackage[bookmarks=true]{hyperref}
\usepackage{tikz}
\usetikzlibrary{shapes,calc,positioning,arrows,}
\pgfarrowsdeclarecombine{twotriang}{twotriang}%
{stealth'}{stealth'}{stealth'}{stealth'} 
\IEEEoverridecommandlockouts
\usepackage{graphicx}
\usepackage{subfig} %
\usepackage[skip=0pt,font=small,labelfont=bf]{caption} %
\usepackage{textcomp}
\usepackage{color}
\usepackage{mathtools}
\usepackage{amsmath}
\DeclarePairedDelimiter{\ceil}{\lceil}{\rceil}
\usepackage{epsfig,graphicx,amsmath,amsfonts}
\usepackage{xcolor,import}
\usepackage{transparent}
\usepackage{upgreek}
\DeclareMathOperator*{\argmin}{arg\,min}
\newcommand\numberthis{\addtocounter{equation}{1}\tag{\theequation}}
\usepackage[ruled,linesnumbered]{algorithm2e}
\usepackage[normalem]{ulem}
\usepackage{multirow}
 
\usepackage{float}
\usepackage[utf8]{inputenc}
\usepackage[TS1,T1]{fontenc}
\usepackage{array, booktabs}
\usepackage{graphicx}
\usepackage{comment}
\usepackage{colortbl}
\usepackage{caption}
\usepackage{amsthm}
\usepackage{cuted}
\usepackage{adjustbox}
\usepackage{pgfplots}
\pgfplotsset{major grid style={dotted,green!50!black}}
\usetikzlibrary{patterns}

\newcommand{\shrinka}{\def\baselinestretch{0.993}\large\normalsize}

\title{\vspace{1em}\LARGE Robust Learning of Tactile Force Estimation through Robot Interaction}
\author{Balakumar Sundaralingam$^{1,2}$ \qquad Alexander (Sasha) Lambert$^{1,3}$ \qquad Ankur Handa$^{1}$ \\[0.1em] \qquad Byron Boots$^{1,3}$ \qquad  Tucker Hermans$^{2}$ \qquad
Stan Birchfield$^{1}$ \qquad Nathan Ratliff$^{1}$ \qquad Dieter Fox$^{1,4}$
\thanks{$^{1}$ NVIDIA, USA.}

\thanks{$^{2}$ University of Utah Robotics Center and the School of Computing, University of Utah, Salt Lake City, UT,
USA. {\tt\small bala@cs.utah.edu}}
\thanks{$^{3}$ Institute for Robotics and Intelligent Machines, Georgia Institute of Technology, GA,
USA. %
}

\thanks{$^{4}$ University of Washington, Paul G.~Allen School for Comupter Science \& Engineering, Seattle, WA, USA}
}
\definecolor{lightgreen}{RGB}{240,249,232}
\definecolor{darkgreen}{RGB}{123,204,196}

\begin{document}
\setcounter{figure}{1}
\makeatletter
\let\@oldmaketitle\@maketitle%
\renewcommand{\@maketitle}{\@oldmaketitle%
\begin{center}
    \centering     
    \includegraphics[width=0.95\linewidth]
    {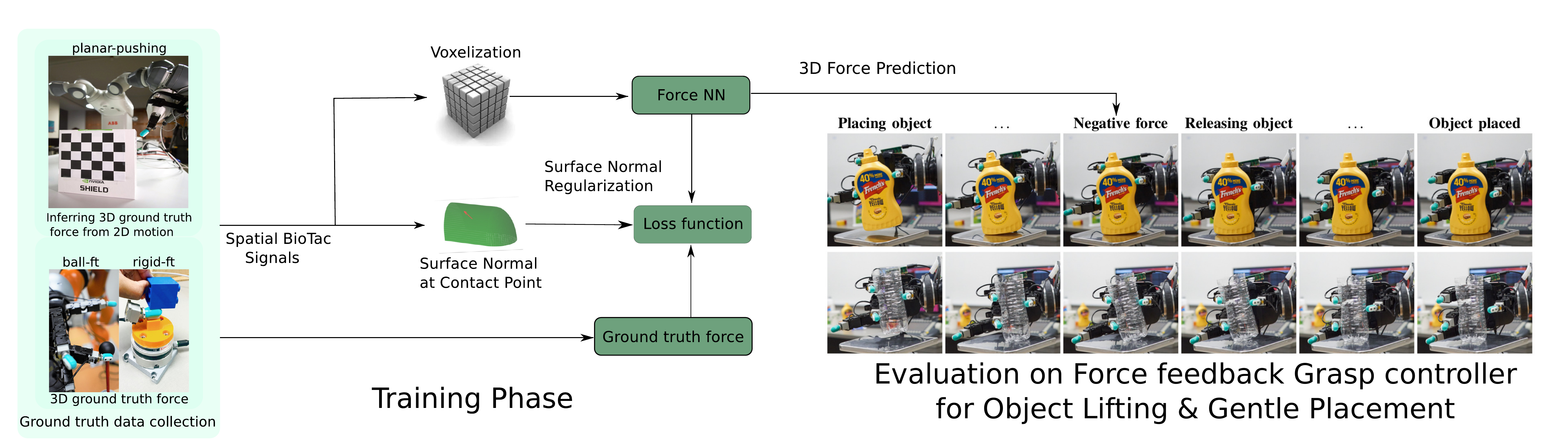}
\end{center}
  \footnotesize{\textbf{Fig.~\thefigure:\label{fig:title}}~The network trained on the datasets collected via three different sources learns to regress the raw tactile sensor readings to the corresponding force vector. The learned force vectors are validated by a force feedback grasp controller performing object lifting and gentle placement.}\vspace{-12pt}
  \medskip}%
\makeatother
\shrinka
\maketitle
\shrinka
\begin{abstract}
Current methods for estimating force from tactile sensor signals are either inaccurate analytic models or task-specific learned models. In this paper, we explore learning a robust model that maps tactile sensor signals to force. We specifically explore learning a mapping for the SynTouch BioTac sensor via neural networks. We propose a voxelized input feature layer for spatial signals and leverage information about the sensor surface to regularize the loss function. To learn a robust tactile force model that transfers across tasks, we generate ground truth data from three different sources:  (1) the BioTac rigidly mounted to a force torque~(FT) sensor, (2) a robot interacting with a ball rigidly attached to the same FT sensor, and (3) through force inference on a planar pushing task by formalizing the mechanics as a system of particles and optimizing over the object motion. A total of 140k samples were collected from the three sources. We achieve a median angular accuracy of 3.5 degrees in predicting force direction~(66\% improvement over the current state of the art) and a median magnitude accuracy of 0.06~N~(93\% improvement) on a test dataset. Additionally, we evaluate the learned force model in a force feedback grasp controller performing object lifting and gentle placement. Our results can be found on \url{https://sites.google.com/view/tactile-force}.
\end{abstract}
\vspace{-1.1em}
\section{Motivation \& Related Work}
\label{sec:introduction}
Tactile perception is an important modality, enabling robots to gain critical information for safe interaction in the physical world~\cite{shenoi2016crf,gao2016deep,calandra2017more}. The advent of sophisticated tactile sensors~\cite{girao2013tactile} with high fidelity signals allows for inferring varied information such as object identity and pose, surface texture, and slip between the object and robot~\cite{xu2013tactile,li2014localization,patel2018integrated,veiga-iros2015-slip-learning,wettels2009grip,meier2016distinguishing,agriomallos2018slippage,heyneman2016slip,hoelscher-ichr2015-tactile-recoginition}. However, using these sensors for force feedback control has been limited to simple incremental controllers conditioned on detection of salient events (\textit{e.g.}, slip or contact)~\cite{veiga2018grip,meier2016distinguishing} or learning task-specific feedback policies on the tactile signals~\cite{sutanto2018learning,chebotar2016self,vanhoof-ichr2015-in-hand-rl}. One limiting factor has been the inaccuracy of functions to map the tactile signals to force robustly across different tasks. Current methods for force estimation on the SynTouch BioTac~\cite{fishel2013syntouch} fail to cover the entire range of forces applied during typical manipulation tasks. Analytic methods~\cite{loeb2013estimating,navarro2015active} tend to produce very noisy estimates at small force values and their accuracy decreases as the imparted force angle relative to the sensor surface normal becomes large (\textit{i.e.}, a large shear component relative to the compression force). On the other hand, learned force models~\cite{wettels2011haptic,su2015force} tend to overfit to the dataset used in training and have not been sufficiently validated in predicting force across varied tasks.

More specifically, Wettel and Loeb~\cite{wettels2011haptic} use machine learning techniques to estimate the force, contact location, and object curvature when a tactile sensor interacts with an object. Lin \textit{et al.}~\cite{loeb2013estimating} improve upon~\cite{wettels2011haptic}, formulating analytic functions for estimation of the contact point, force, and torque from the BioTac sensor readings. Navarro \textit{et al.}~\cite{navarro2015active} explore calibration of the force magnitude estimates by recording the DC pressure signal when the sensor is in contact with a force plate. They use these values in a linear least squares formulation to estimate the gain. While they can estimate the magnitude of force, they cannot estimate force direction. Su \textit{et al}.~\cite{su2015force} explore using feed-forward neural networks to learn a model that maps BioTac signals to force estimates. The neural network more accurately estimates forces than the linear model from~\cite{loeb2013estimating} and is used to perform grasp stabilization. Importantly, none of these methods validate their force estimates using a data source different from the method used to generate the training data. They also lack experimental comparison between different approaches in the context of robotic manipulation tasks.

In this paper, we attempt to address these shortcomings, by collecting a large scale ground truth dataset from different methods and by leveraging the sensor surface and spatial information in our proposed neural network architecture. For one of our collection methods, we infer force from the motion of an object on a planar surface, by formalizing the interaction as a system of particles, a deviation from the well-established velocity model for planar pushing~\cite{lynch1996stable} which does not reason about force magnitude. This scheme of force estimation allows us to obtain accurate small-scale forces~(0.1-2N), enabling us to learn a precise force prediction model.

Motivated by~\cite{reinecke2014experimental}, we compare our proposed method with the current state-of-the-art methods for force estimation for the BioTac sensor. We specifically compare the analytic model from~\cite{loeb2013estimating} and the best performing feed-forward neural network model from~\cite{su2015force}. We compare both in terms of force estimation accuracy on our dataset and also empirical experiments on a robot manipulation task. To summarize, this paper makes the following contributions:
\begin{enumerate}
\item We provide a novel method to infer force from object motion on a planar surface by formalizing the mechanics as a system of particles and solving for the force in a least squares minimization problem, given the object motion and the point on the object where the force is imparted.
\item We introduce a novel 3D voxel grid, neural network encoding of tactile signals enabling the network to better leverage spatial relations in the signal. We further tailor our learning to the tactile sensor through the introduction of a novel loss function used in training that scales the loss as a function of the angular distance between the imparted force and the surface normal.
\item We collected a large-scale dataset for the BioTac sensor, consisting of over 600 pushing episodes and 200 interactions between an arm-hand system equipped with the BioTac sensors and a force torque sensor.
\end{enumerate}
We validate these contributions on our dataset and in an autonomous pick and place task. We show that our proposed method robustly learns a model to estimate forces from the BioTac tactile signals that generalize across multiple robot tasks. Our method improves upon the state of the art~\cite{su2015force,loeb2013estimating} in tactile force estimation for the BioTac sensor achieving a median angular accuracy of 3.5~degrees in predicting force direction~(66\% improvement over the current state of the art) and a median magnitude accuracy of 0.06~N~(93\% improvement) on a test dataset.

\section{Problem Definition \& Proposed Approach}
\label{sec:problem-definition}
We describe the sensor's states in the following section, followed by a formal definition of the problem. We then describe the computation of ground truth force from planar pushing in Sec.~\ref{sec:phys_pushing} and our network architecture in Sec.~\ref{sec:network-architecture}.
\subsection{BioTac Sensor}
\label{sec:biotac-sensor}
We use the BioTac sensor~\cite{wettels2014multimodal} from SynTouch. The sensor has a rigid core which is enveloped by a high friction elastomeric skin. A weakly conductive liquid is filled in the space between the core and the skin. There are 19 impedance sensing electrodes spread out on the core surface, giving measurements~$e\in\mathbb{R}^{19}$. A thermistor coupled with heaters measures the fluid temperature~$T_{dc}\in\mathbb{R}$ and temperature flow~$T_{ac}\in\mathbb{R}$.  A transducer measures the static pressure~$p_{dc}\in\mathbb{R}$. High frequency changes to the pressure are measured by the transducer at 2.2~kHz and sent to the system in a buffer with the past 22 values~$p_{ac}\in\mathbb{R}^{22}$ at 100~Hz along with the other signals. 
A single sensor sample is thus given by~$z=[e^\top,p_{dc},p_{ac}^\top,T_{dc},T_{ac}]^\top\in\mathbb{R}^{44}$.
Following~\cite{loeb2013estimating}, we use the tared 
signals from the sensor (\textit{i.e.}, initial value subtracted). 
Using methods described by Lin et al.~\cite{loeb2013estimating}, we also compute the contact point~$s_c\in\mathbb{R}^3$ on the BioTac sensor and the surface normal~$s_n\in\mathbb{R}^3$ by approximating the BioTac surface geometry as a half-cylinder attached to a quarter-cylinder cap, both of the same radius~$r$.
\vspace{-0.5em}
\subsection{Problem Definition \& Approach Overview}
We define the problem as estimating the force~$f\in\mathbb{R}^3$ with reference to the sensor frame~$B$, given~$z$, $s_c$, and $s_n$. We use feed-forward neural networks to learn the function $f=F(z,s_c,s_n)$ that maps from sensor readings $z$, the sensor surface contact point~$s_c$, and the surface normal~$s_n$, to the force $f$. In order to learn an accurate model, our training dataset needs to cover a wide range of forces~(in magnitude and direction). Furthermore, to learn a robust model that transfers to new tasks, we generate ground truth data from three different sources. The first source is collected by rigidly attaching the BioTac to a wrist force/torque~(FT) sensor similar to~\cite{su2015force,loeb2013estimating} and pressing on the BioTac sensor using objects. We term this source~\emph{rigid-ft}.  This requires a human to interact with the object and is biased by the human. This setup was used to cover very large forces. For the second source, we attach the same FT sensor to a ball, with which we interact using a robotic hand-arm system. We call this source~\emph{ball-ft}. The \emph{ball-ft} source adds randomness to the orientation of the BioTac sensor frame with respect to the force torque sensor frame. The wrist force/torque sensor is noisy in ranges between 0.01~N to 0.1~N, making small force readings unreliable. To overcome this problem, we collect sensor readings from a planar pushing setup, where a robot pushes a box on a planar surface using the tactile sensor. We call this source~\emph{planar-pushing}. The ground truth force for planar pushing is computed by least squares optimization, described in the next section.
\begin{figure*}[h]
  \centering
  \includegraphics[width=0.83\textwidth]{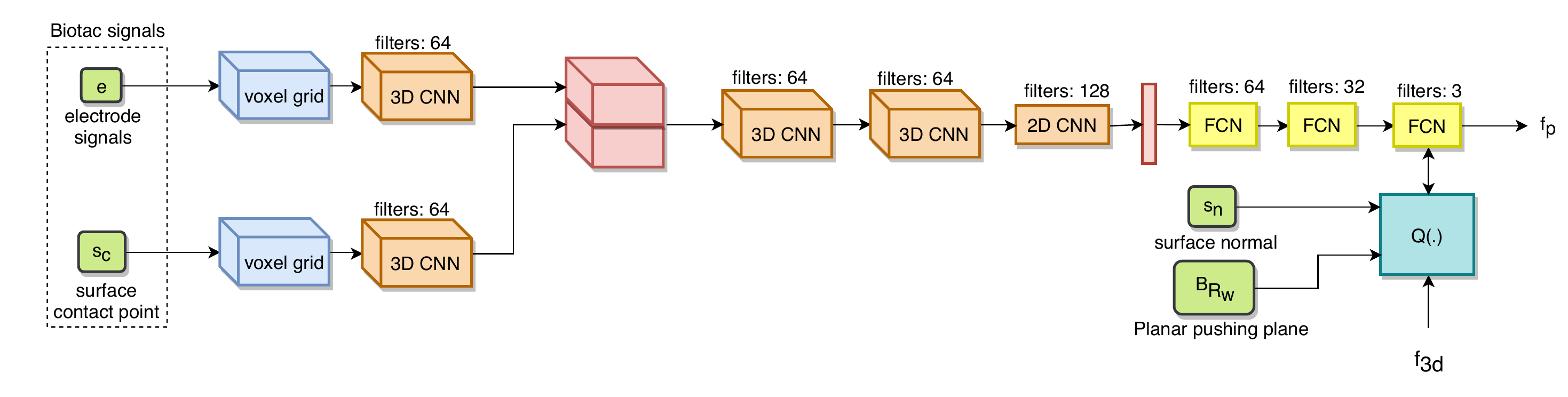}
  \caption{The force prediction neural network uses 3D voxelized inputs that preserve the spatial information. We use layer norm followed by ReLU after every convolutional and fully connected layer (FCN). Additionally, we use kernels and strides of 2 for every convolutional layer.}
  \label{fig:force_nn}
\end{figure*}
\vspace{-0.4em}
\subsection{Mechanics of Planar Pushing as a System of Particles}
\label{sec:phys_pushing}
Given an object with mass~$m$ in an SE(2) planar space, moving with a linear velocity~$v$ and an angular velocity~$\omega$, the net force causing this motion can be defined as
\begin{align*}
  f_c&=m\dot{v} \numberthis \label{eq:linear_acc}\\
  c \times f_c&=I\dot{\omega} \numberthis \label{eq:angular_acc}
\end{align*}
where $f_c\in\mathbb{R}^2$ is the net force acting at a point~$c\in \mathbb{R}^2$ with reference to the center of mass~(CM) of the object. Given the linear acceleration, the net force can be obtained. However, if the measurement system is not able to observe small linear accelerations, solving Eq.~(\ref{eq:linear_acc}) is intractable. There are two cases when the linear acceleration can be small: 1) when the force applied to the object is very small, causing very small linear and angular acceleration, 2) when the force applied is perpendicular to the radial line, in which case the object will have a large angular acceleration. In the latter case, Eq.~(\ref{eq:angular_acc}) could give us the net force~$f_c$. However, Eq.~(\ref{eq:angular_acc}) is a degenerate system as we need~$f_c\in\mathbb{R}^2$ from~$\dot{\omega}\in\mathbb{R}$. We solve for~$f_c$ by formulating Eq.~(\ref{eq:linear_acc}) and Eq.~(\ref{eq:angular_acc}) as loss functions in a least squares minimization problem:
\begin{align*}
  \argmin_{f_c} \quad & k||f_c-m\dot{v}||^2 + ||[c]_\times f_c-I\dot{\omega}||^2 \numberthis \label{eq:force_opt}
\end{align*}
where the weight~$k$ scales the linear acceleration loss and $[c]_\times$ is the skew symmetric matrix of vector~$c$.

Consider the object to be now resting on a planar surface with coefficient of friction~$\mu_s$ between the object and the surface. The friction between the object and the planar surface will oppose the motion of the object with a frictional force~$f_f$ and moment~$n_f$. If the contact region between the object and the surface is~$R$, and $r$ is any point on the object in this region, the force and moment can be defined using Coulomb's law as 
\begin{align*}
  f_f&=-\mu_s \int_{R} \frac{v(r)}{\|v(r)\|}p(r)dA\numberthis\label{eq:friction_trans_int}\\
  n_f&=-\mu_s \int_{R} r \times \frac{v(r)}{\|v(r)\|}p(r)dA \numberthis\label{eq:friction_moment_int}
\end{align*}
where $v(\cdot)$ is a function that gives the velocity of the point. The pressure at $r$ is given by~$p(\cdot)$ and $dA$ is a differential element of area at $r$. We derive the moment with reference to the object's center of mass.
To make computation of the frictional force tractable for planar pushing, we make the following assumptions:
\begin{enumerate}
\item The pressure distribution in the contact region~$R$ is uniform.
\item The rigid body is made of~$n$ particles which are uniformly distributed.
\end{enumerate}
The contact region is decomposed into~$n$ small regions, with center of mass for region~$i$ at~$r_i$ and the normal force applied by region~$i$ is~$\int_ip(r_i)dA={\frac{mg}{N}}$.

With the listed assumptions, we simplify the frictional force 
\begin{align*}
  f_f&=-\mu_s {\frac{mg}{N}}\sum_{i=0}^n \frac{v(r_i)}{\|v(r_i)\|}\numberthis \label{eq:friction_force}
\end{align*}
and the moment due to frictional force becomes
\begin{align*} 
  n_f&=-\mu_s {\frac{mg}{N}} \sum_{i=0}^n r_i \times \frac{v(r_i)}{\|v(r_i)\|}\numberthis \label{eq:friction_moment}
\end{align*}

Including the frictional force~$f_f$ and moment~$n_f$ in our minimization problem of~Eq.~(\ref{eq:force_opt}),
\begin{align*}
  \argmin_{f_c} \quad & k||f_c+f_f-m\dot{v}||^2 + ||[c]_\times f_c +n_f-I\dot{\omega}||^2 \numberthis \label{eq:push_force_opt}
\end{align*}
Optimizing~Eq.~(\ref{eq:push_force_opt}) yields an estimate of the force~$f_c$. The force~$f_c$ is 2D, parallel to the support surface. We obtain the ground truth force~$f_{3d}=\prescript{B}{}{R}_o f_c $ by transforming the force~$f_c$ from the object's frame of reference~$o$ to the BioTac sensor frame~$B$.
\subsection{Network Architecture}
\label{sec:network-architecture}
Our proposed neural network architecture takes only the spatial signals\footnote{We found empirically that the other signals did not improve the force estimation accuracy.}~$e,s_c$ from the BioTac to estimate the force as shown in Fig.~\ref{fig:force_nn}. We create a 3D voxel grid and input the value of each electrode on the corresponding voxels based on the electrode's position with reference to the BioTac frame~$B$. We create a second voxel grid for the contact point and input a value of 1 for the voxel at contact point~$s_c$. These two voxel layers are concatenated and passed through two layers of 3D convolutions. The features are then flattened and passed through a layer of 2D convolutions, which is further flattened to a vector. This vector passes through fully connected layers to output the predicted force vector $f_p$ of length 3. 

\subsection{Loss Functions}
The predicted force vector~$f_p$ is %
compared to the 3D ground truth force~$f_{3d}$ via a scaled $\ell_2$ norm.
\begin{align*}
  Q_{3d}(f_{3d},f_p)&= \frac{1}{||f_{3d}||}||f_{3d}- f_p|| \numberthis\label{eq:3d_loss}
\end{align*}

For the planar pushing dataset, we use a projected~$\ell_2$ norm, as there could be forces acting perpendicular to the planar surface which the physics model does not take into account.
\begin{align*}
  Q_{proj}(f_{3d},f_p,\prescript{w}{}R_B)&= \frac{1}{||f_{3d}||} ||(\prescript{w}{}R_B  f_{3d}-\prescript{w}{}R_B f_p)^\top\psi||^2 \numberthis\label{eq:2d_loss}
\end{align*}
where $\psi$ is the orientation of the support surface plane.

The high friction of the BioTac surface allows for imparting force from directions other than the surface normal at a contact point. So the force could be applied from any contact point on the surface and is not only limited to contact points whose surface normal matches with the force direction. We hypothesize that as the angle between the force and the surface normal increases, the sensor's signals might be less meaningful. We scale the loss function  with an adaptive weight function~$\alpha(\cdot)$ to reflect this hypothesis. We empirically analyze this hypothesis in Sec.~\ref{sec:results}.
\begin{align*}
  \alpha(s_n,f_{3d})&=2^{\beta(1-D(s_n,f_{3d}))} 
  \numberthis\label{eq:alpha_weight}\\
  D(s_n,f_{3d})&=\frac{\cos^{-1}(s_n^\top  \hat{f}_{3d})}{\pi} \numberthis\label{eq:cos_distance}
\end{align*}
where $\beta$ is a scalar weight and~$\hat{f}_{3d}$ is the unit vector of the ground truth force vector~$f_{3d}$ and $D(\cdot)$ is the normalized cosine distance function. The loss function used in our network is defined as
\begin{align*}
  Q(\cdot)&=\begin{cases}
    \alpha(\cdot)Q_{proj}(f_{3d},f_p,\prescript{w}{}R_B) & \text{if planar pushing}    \\
    \alpha(\cdot)Q_{3d}(f_{3d},f_p) & \text{otherwise}
    \end{cases}
\end{align*}

\section{Dataset Collection, Implementation Details, \& Experimental Protocol}
\label{sec:impl-det}
This section provides a concise description of our dataset collection procedure. We also provide implementation details of our neural network and describe the error metrics and comparison methods used to analyze our force model.
\subsection{Dataset Collection}
The setups for our three different data sources~\emph{rigid-ft}, \emph{ball-ft}, and \emph{planar-pushing} are shown in Fig.~1. For learning the force model, we only used samples from the dataset that have non-zero force readings; we term these samples "force samples". We use the OptoForce HEX-E 6-DOF force torque sensor to collect the \emph{rigid-ft} and \emph{ball-ft} data. For \emph{rigid-ft}, we mounted the BioTac to the force torque~(FT) sensor and pressed down on the finger using flat rigid plastic objects to collect data. This closely resembles the data collection performed by~\cite{loeb2013estimating,su2015force} for BioTac force estimation. We collected a total of 20k force samples. For the \emph{ball-ft} method the Allegro hand pushed the BioTac against a hard plastic ball mounted on a vertical bar attached to the same FT sensor. We generated a total of 200 random trajectories for the middle fingertip to make contact with the ball across 10 different wrist poses generating a total of 20k force samples. For the \emph{planar-pushing} method, we mounted a BioTac on an ABB YuMi robot which pushed a known box weighing 0.65~kg. We generated a single straight-line, task-space position trajectory for the YuMi fingertip to follow using trajectory optimization. We use Riemannian motion policies~\cite{Cheng-WAFR-18} to execute the task space trajectory. We chose a random initial orientation for the BioTac and box for every execution of the task space position trajectory. The orientation of the box was sampled from a small range to keep the contact on the same face of the box for each push. We collected a total of 600 trials on the robot generating 100k force samples in total. 

Our final dataset collected across all three data sources contains a total of 140k force samples. For \emph{ball-ft} and \emph{planar-pushing} setups, we track the robot with an ASUS Xtion RGB-D camera using Dense Articulated Real-Time Tracking~(DART)~\cite{schmidt2015depth}. We enable DART's contact prior when the FT sensor measures a force greater than 2~N for \emph{ball-ft} and when the BioTac absolute pressure signal~($p_{dc}$) rises greater than 10 units for the \emph{planar-pushing} setup.

We chose the parameters for the optimization described in Eq.~(\ref{eq:push_force_opt}) as $n=80$ and $k=10$. We found the number of particles~$n$ did not affect the force by much above this size and any value of~$k$ greater than 2 gave similar performance. We chose the coefficient of friction~$\mu_s=0.1$ between the box and the planar surface by interpolating data from~\cite{friction_url}. We solve the optimization using Sequential Least Squares Programming~(SLSQP)~\cite{kraft1988software} available through PAGMO~\cite{biscani2010global}.
\subsection{Neural Network Implementation Details}
We built our neural network in TensorFlow~\cite{abadi2016tensorflow}. For each data source, we used 80\% for training and approximately 10\% for validation and from the remaining data, we picked 1.5k samples for testing; the data was split by trials~(leaving whole trials). We run the training for a maximum of 200 epochs with a batch size of 512 and store the model only when the loss on the validation set improves. We optimize using the Adam optimizer~\cite{kingma2014adam}. We use an adaptive learning rate that starts at \(10^{-4}\) and increases for the first 2 epochs by~$2^{\ceil{(i / 50)}}$ and later decreases by 0.95 each iteration~$i$ for the remaining epochs, motivated by~\cite{smith2017super}. 

We only send the BioTac signals to the network when the robot detects the fingertip is in contact. We use the absolute pressure ($p_{dc}$) signal from the BioTac to determine contact and classify the sensor as in contact if this $p_{dc}$ signal maintains a value above 10 for the past 10 timesteps. This reduces false positives when the sensor is moving in free space. We set the voxel grid to have dimensions $15 \times 15 \times 7$, which allowed a unique voxel for each electrode of the BioTac sensor. We plan on studying the effect of voxel dimensions on the efficiency of learning in the future.
\subsection{Error Metrics, Protocol \& Comparison Methods}
Given the predicted force~$f_p$ and the ground truth force~$f_{3d}$, we compute the error in direction as the cosine similarity between the vectors. We scale this cosine similarity to give a percentage direction accuracy.
\begin{align*}
    \text{Direction error\%}&=100\times \frac{1}{\pi}\cos^{-1}(\hat{f}_{3d}^\top \hat{f}_{p}) \numberthis \label{eq:direction_acc}
\end{align*}
For computing the error in magnitude, we report the symmetric mean absolute percentage accuracy between the ground truth and predicted force magnitudes:
\begin{align*}
    \text{Magnitude error\%}&=100\times \frac{\text{abs}(||f_{3d}|| -||f_p||)}{||f_{3d}|| + ||f_p||}\numberthis \label{eq:mag_acc}
\end{align*}
We report the absolute $\ell_1$ norm between the ground truth and predicted force magnitude as "magnitude error~(N)".

We compare our proposed network architecture against the NN-3 model from Su et al.~\cite{su2015force} and also the linear model given by~\cite{loeb2013estimating}:
\begin{align*}
    f_p & = [S_x \sum_{i=1}^{19}e_i n_{x,i}, S_y \sum_{i=1}^{19}e_i n_{y,i}, S_z \sum_{i=1}^{19}e_i n_{z,i}]^\top \numberthis \label{eq:analytic_model}
\end{align*} We compute the scalar weights~$S_x,S_y,S_z$ using linear regression and $[n_{x,i},n_{y,i},n_{z,i}]^\top$ for the orientation of electrode~$i$.

We report error on the test dataset which contains 1500 samples from each data source~(4500 in total). To study how each dataset source affects the prediction model, we train each prediction model with the following source combinations:
\begin{enumerate}
    \item \emph{rigid-ft} to directly compare to Su et al.~\cite{su2015force} and~\cite{loeb2013estimating}.
    \item \emph{planar-pushing} to study how 2D ground truth force obtained by optimizing over planar physics performs on predicting 3D forces.%
    \item \emph{ball-ft} to study the effect of randomization of the sensor's frame with respect to a 6-DOF force sensor.
    \item \emph{mixed}, which includes all sources~\emph{rigid-ft}, \emph{planar-pushing} and \emph{ball-ft}, to study the ability of the prediction model to learn from different sources.
\end{enumerate}

We also evaluate the effect of the two primary contributions of our network structure: spatial signal encoding by 3D voxelization and the proposed~$\alpha$ weight. We train our proposed network architecture with and without each of these contributions~(making a total of four models) on the same source combinations as described above. We compare voxelization to  four fully connected layers of the signals~$e$ and~$s_c$, input to the proposed network's first fully connected layer. 
\begin{figure}
    \centering
    \includegraphics[trim={0cm 3.2cm 0cm 0cm},clip,width=0.9\columnwidth]{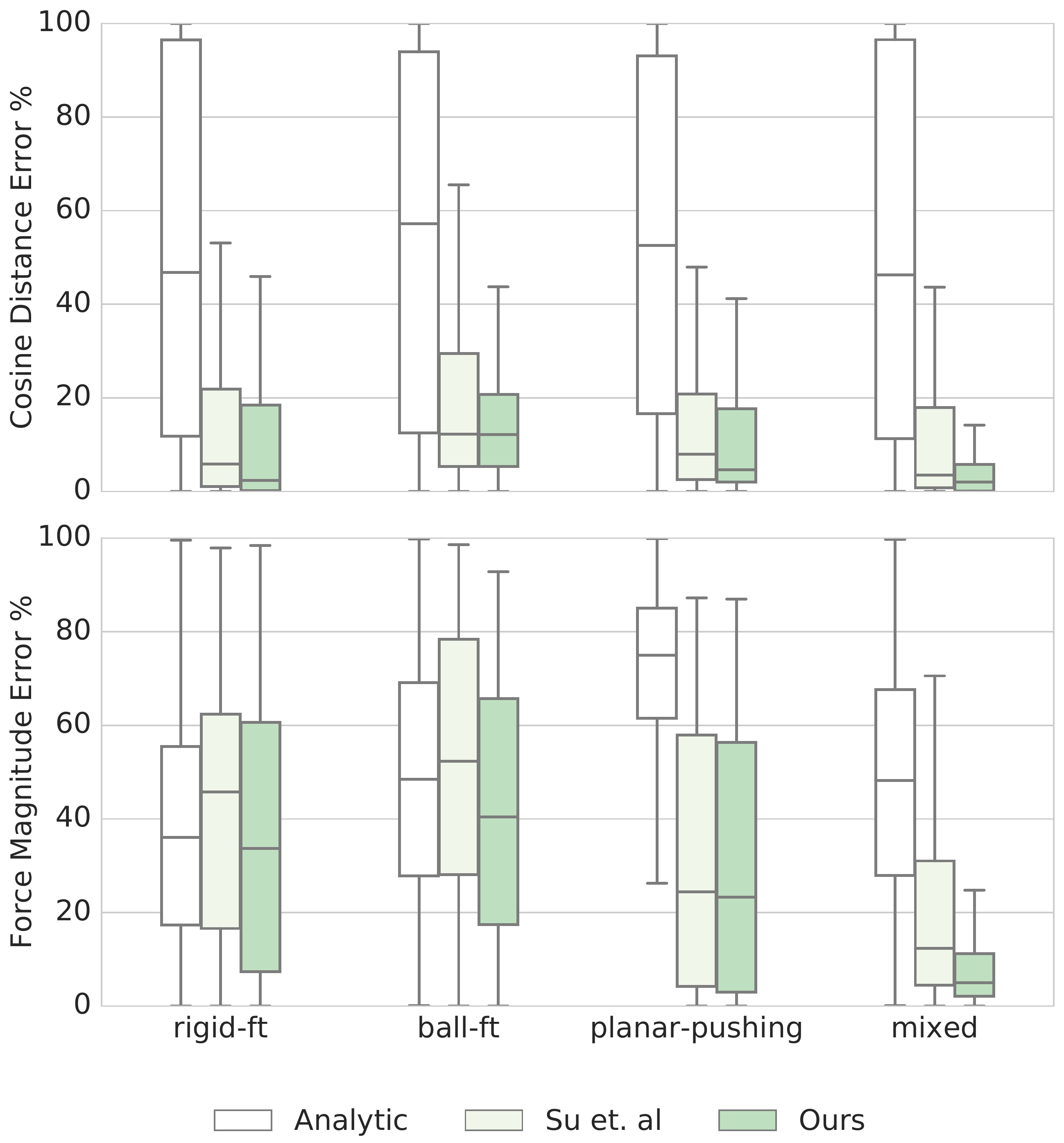}
    \includegraphics[trim={0cm 0cm 0cm 12.35cm},clip,width=0.9\columnwidth]{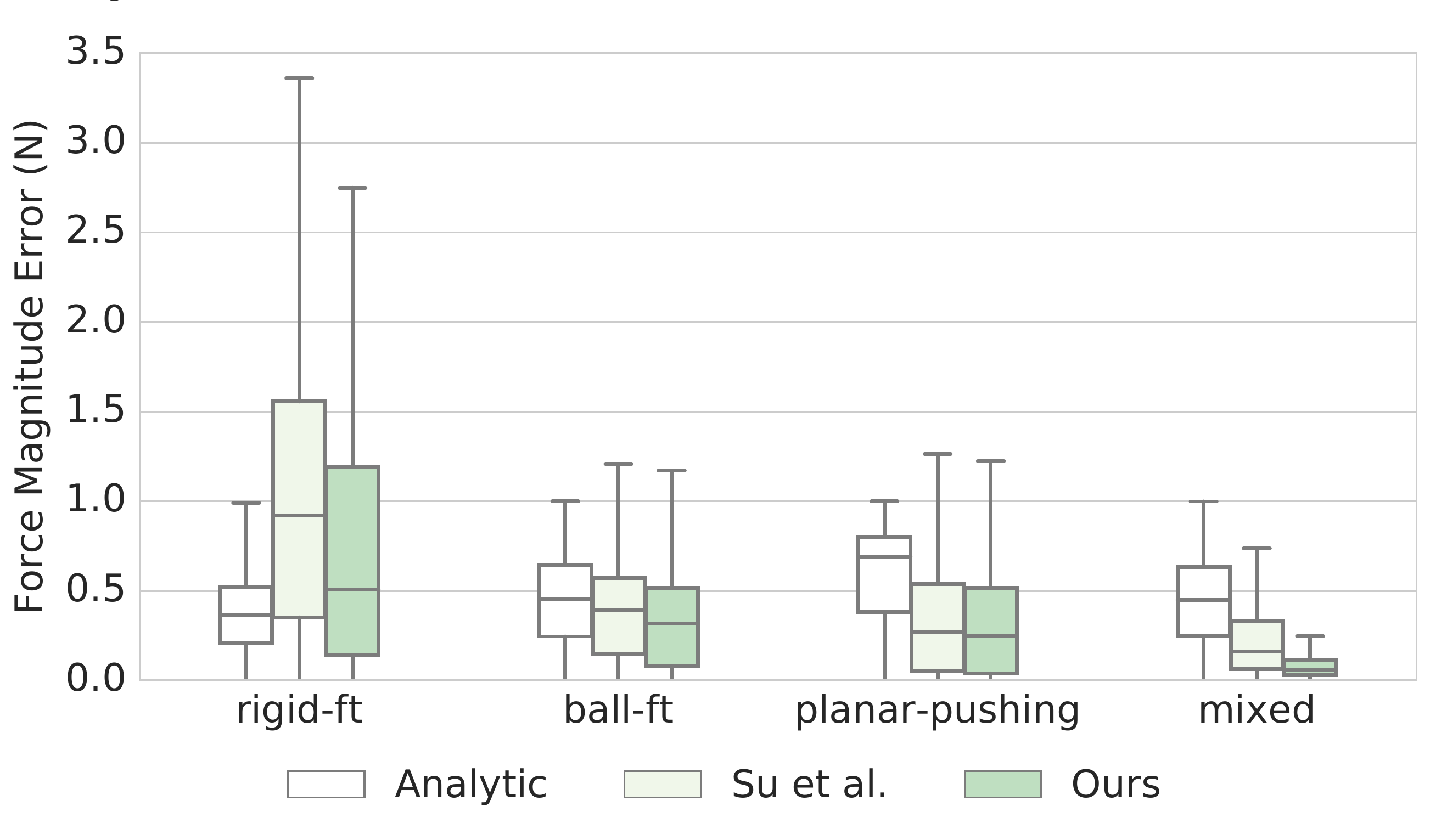}
    \caption{Predicted force error for different models and training sets. Analytic refers to the linear model from~\cite{loeb2013estimating}, Su \textit{et al.} refers to the best performing model from~\cite{su2015force}.}
    \label{fig:plot_methods}
\end{figure}

\section{Results}
\label{sec:results}
We now report the results on our test dataset. In all plots the middle line in the box plot defines the median error. The bottom and top borders indicate the first and third quartiles. The whiskers indicate the extrema of the inliers within 1.5 times the interquartile range.

\begin{figure}[t]
  \centering
  \includegraphics[width=0.85\columnwidth]{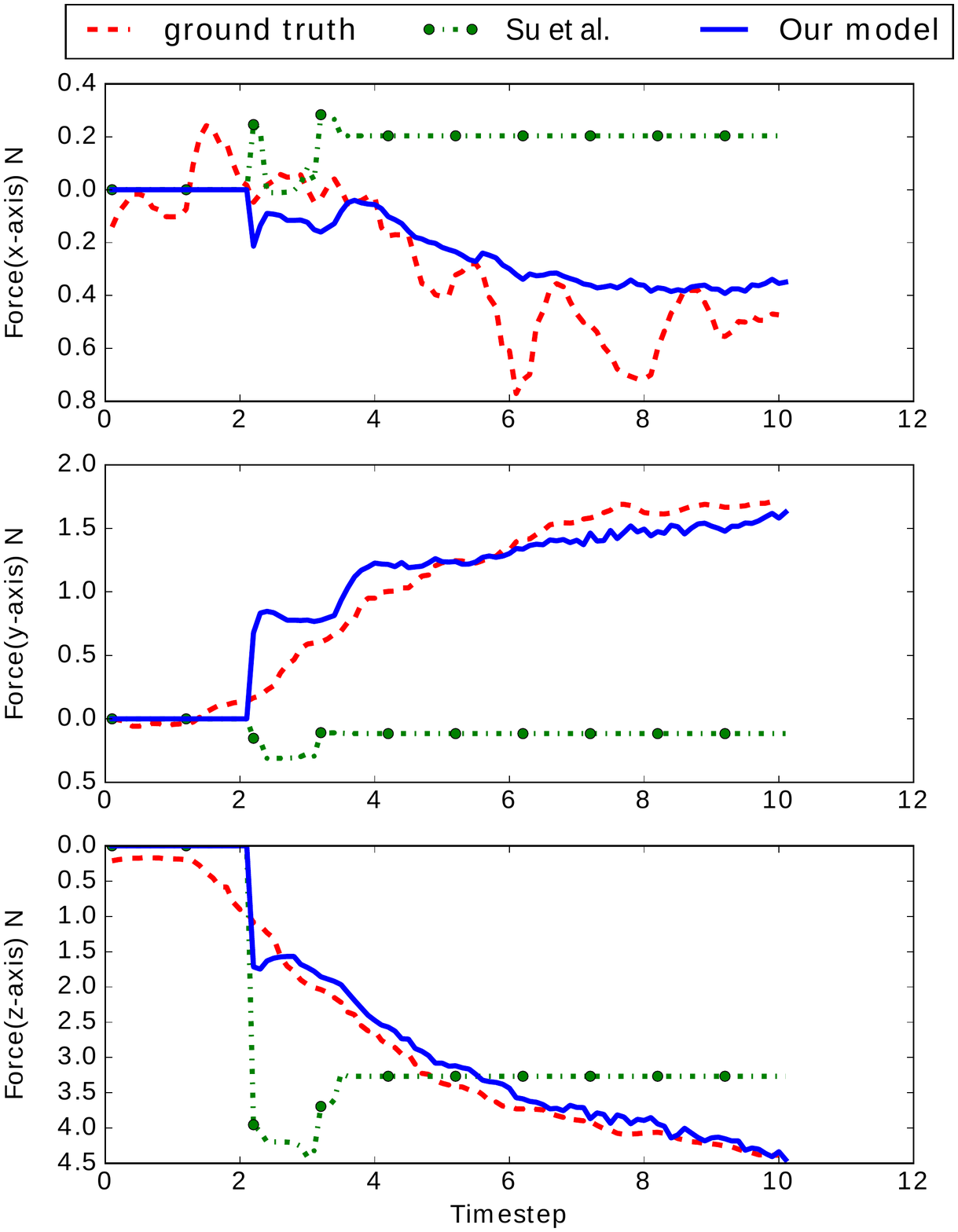}
  \caption{Estimated force from our model and Su~\textit{et al.} compared with the FT sensor, to which the BioTac is rigidly mounted. Our model sufficiently tracks the ground truth along all three axes.}%
  \label{fig:plot_force_t}
\end{figure}

\noindent\textbf{Prediction accuracy:}
Our method trained on the \emph{mixed} dataset achieves the best accuracy, as shown in~Fig.~\ref{fig:plot_methods}. We achieve a median angular error of 0.06 radians~(3.5 degrees) and a median magnitude error of 0.06 N compared to Su \textit{et al.}'s median angular error of 0.18 radians~(10 degrees) and median magnitude error of 0.91 N.  We also compare our model to Su~\textit{et al.}~(trained on \emph{rigid-ft}) in a time series of force estimates in Fig.~\ref{fig:plot_force_t}. We see that our model sufficiently captures the ground truth while Su~\textit{et al.} only covers the magnitude along $z$-axis. In Fig.~\ref{fig:plot_force_t}, we suspect the FT ground truth to be noisy, causing the oscillating behaviour in the force along $x$-axis.

\begin{figure}[t]
  \centering
  \includegraphics[width=0.8\columnwidth]{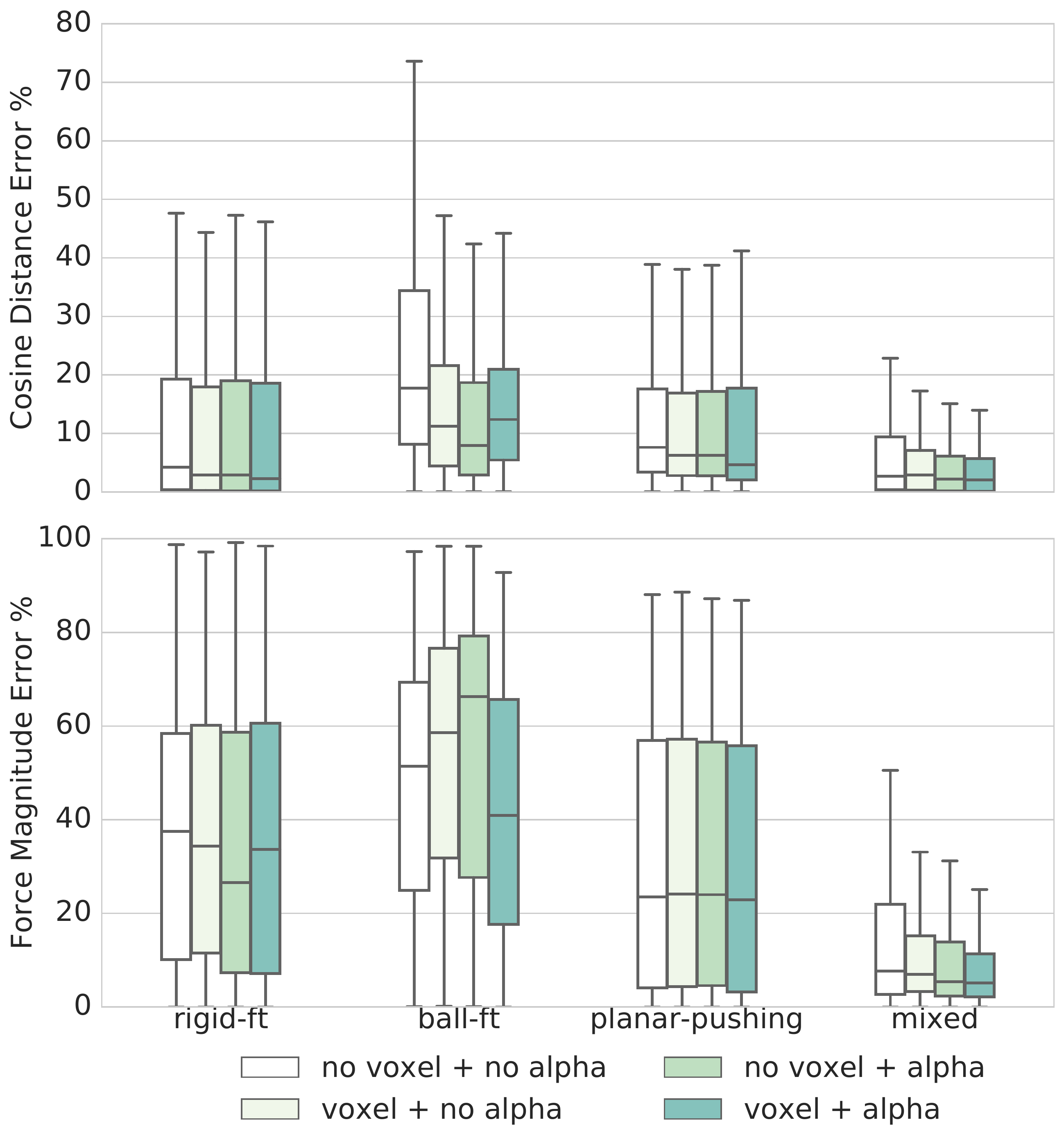}
  \caption{Effect of the spatial encoding ("voxel") and $\alpha$ on the prediction accuracy.}
  \label{fig:plot_network_alpha}
\end{figure}

\noindent\textbf{Effect of spatial encoding and $\alpha$ regularization:}
We investigate the effect of $\alpha$ in the loss function. As seen in Fig.~\ref{fig:plot_network_alpha}, without~$\alpha$, the interquartile is larger, specifically in~\emph{ball-ft} and \emph{mixed} trained models, indicating that~$\alpha$ indeed helps regularize the force predictions. 
Voxelization helps in lowering the prediction error further when combined with~$\alpha$ highlighting that spatial information is important. Without~$\alpha$, voxelization performs worse only on the magnitude of \emph{ball-ft} trained model.

\vspace{-5pt}
\section{Force Feedback for Object Manipulation}
\vspace{-3pt}
We analyze the generalization ability of the learned force model in an autonomous object lifting and placement task. We use this task to illustrate the utility of the learned force model and do not directly compare to other approaches to grasp stabilization and placement using tactile sensing~\cite{romano2011human,veiga-iros2015-slip-learning}. 

To perform the task the robot reaches its hand to a fixed pose with respect to the object. The robot closes its fingers and makes contact in the desired configuration. The robot then attempts to increase the force on all the fingertips to a desired threshold~$\tau$ by increasing the finger position along a predefined task-space vector. We experimentally selected $\tau$ to be 2~N as the minimal force needed for the method from~\cite{su2015force} to lift the \emph{soft-scrub} object. The robot then raises its arm to lift the object along a straight-vertical trajectory. After reaching the desired height, the robot lowers its arm down along the same straight line and the fingers release the object when a negative force in the normal direction to the support surface is detected on the index fingertip. All grasps we studied were three fingered grasps. We chose the index finger as it had the largest change in force direction caused by the support surface. 

The lifting and placement tasks directly depend on the accuracy of the forces estimated from the tactile sensor. If the estimated force values are larger than the actual force, the object will not be in a stable grasp and the object will not rise with the arm. If the values are smaller than the actual force, the grasp may deform the softer objects used in the experiments. If the force estimates are incorrect at placement, then the robot will either drop the object too soon or push down onto the table with excessive force, possibly knocking the object over. 

\begin{figure}[t]
   \centering
   \begin{adjustbox}{minipage={\columnwidth}}
    \begin{tikzpicture}
      \begin{axis}[
        ybar, ymin=-1, ymax=100,
        ylabel={Success Percentage (\%)},
        axis lines*=left,
        y label style={at={(axis description cs:0.045,0.5)},anchor=north},
        symbolic x coords={pringles, soft-scrub, lego, plastic-bottle, paper-cup, mustard, all},
        xtick=data,
        x tick label style={font=\scriptsize,rotate=0,yshift=0.0em,text width=0.8cm,align=center},
        xticklabels={chips, cleanser, lego, plastic-bottle, paper-cup, mustard, all},
        extra x tick style={%
    draw=none,fill=none,tick label style={rotate=0,draw=none,yshift=-1.5em,xshift=0.0em}
  },
  extra x ticks={pringles, soft-scrub, lego, plastic-bottle, paper-cup, mustard, all},
  extra x tick labels={{\includegraphics[trim={10cm 2cm 26cm 2cm},clip,height=2cm]{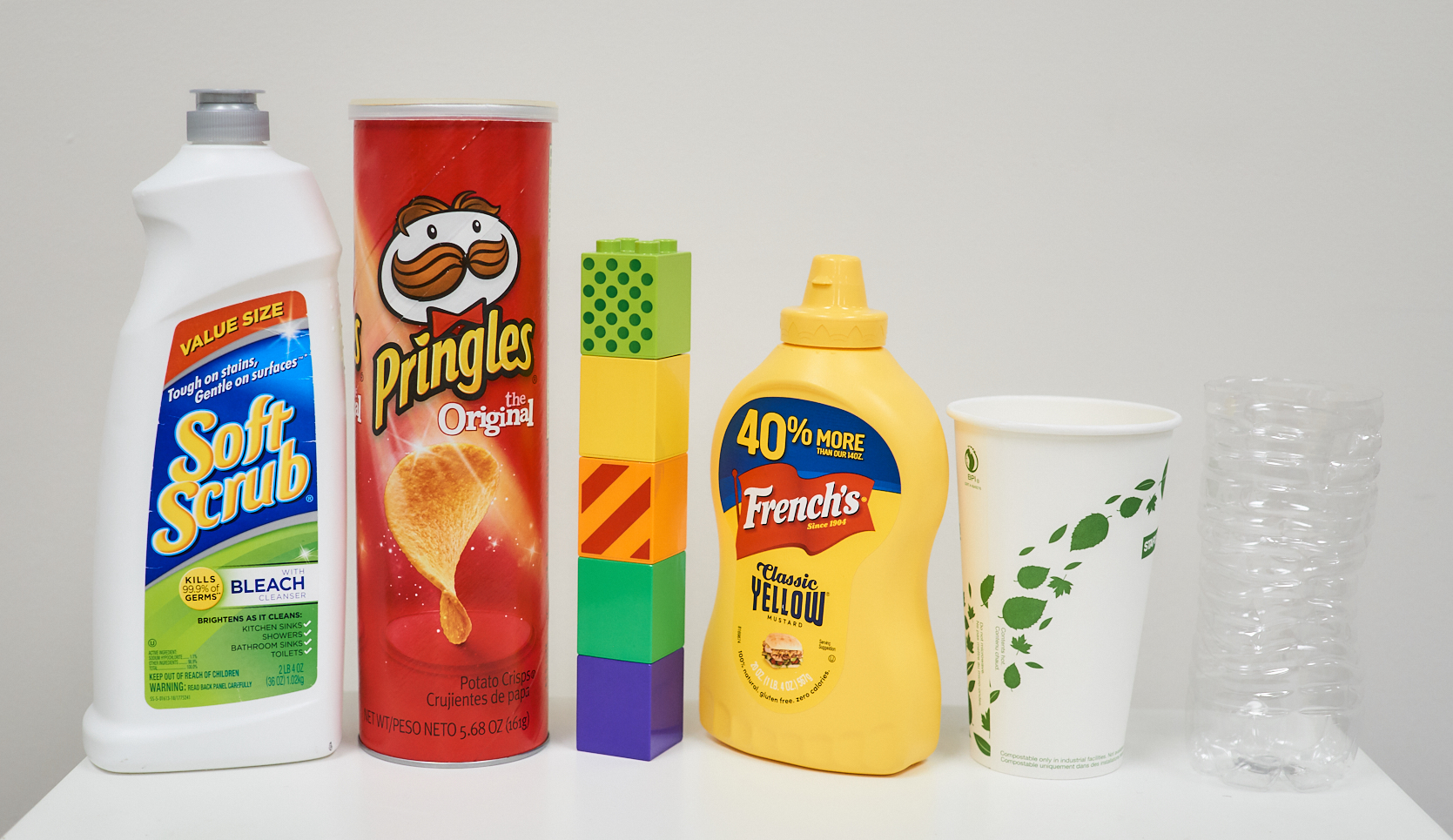}},{ \includegraphics[trim={2cm 2cm 32.4cm 2cm},clip,height=2cm]{figs/object}}, {\includegraphics[trim={16.3cm 2cm 21.8cm 2cm},clip,height=2cm]{figs/object}},{\includegraphics[trim={34.5cm 1.2cm 1cm 2cm},clip,height=2cm]{figs/object}}, {\includegraphics[trim={27.8cm 1.2cm 7.5cm 2cm},clip,height=2cm]{figs/object}}, {\includegraphics[trim={20.3cm 2cm 14.5cm 2cm},clip,height=2cm]{figs/object}},},
  ymajorgrids=true,
        ticklabel style = {font=\scriptsize},
        legend style={font=\scriptsize, draw=none, fill=none},
        ylabel style = {font=\scriptsize},
        bar width = 7pt, height=3.5 cm, width=1.08\columnwidth,
        legend style={area legend, at={(1,1.25)}, anchor=north east, legend columns=2, },
        legend image code/.code={%
         \draw[#1] (0cm,-0.1cm) rectangle (2mm,1mm);},
        ]
        \addplot [fill=lightgreen] coordinates {
          (pringles, 100) (soft-scrub, 100) (lego, 100) (plastic-bottle, 40) (paper-cup, 0)
          (mustard, 20) (all, 60)};
        \addplot [fill=darkgreen] coordinates {
          (pringles, 100) (soft-scrub, 100) (lego, 80) (plastic-bottle, 100) (paper-cup, 80)
          (mustard, 80) (all, 90)};
        \legend{Su et al.,Ours}
      \end{axis}
    \end{tikzpicture}
  \end{adjustbox}
    \caption{Success rates on the manipulation task of object lifting and placement. Our method performs significantly better on the deformable objects~\emph{plastic-bottle} and~\emph{paper-cup}.}
  \label{fig:plot_placement}
 \end{figure}

We chose four objects from the YCB dataset~\cite{calli2015} and two deformable objects---a paper cup and a plastic bottle. The objects are shown in Fig.~\ref{fig:plot_placement}. We show results for our \emph{mixed} dataset trained prediction model and compare to Su et al.'s prediction model~\cite{su2015force}. We train the Su et al. prediction model using the \emph{rigid-ft} data to closely replicate the experiments in~\cite{su2015force} and show the benefit of training on diverse sources of data. We ran 5 manipulation trials per object for each method. The initial and desired pose of the object was kept consistent across the two methods.

On rigid objects, both methods performed similarly well, as shown in Fig.~\ref{fig:plot_placement}. However, on deformable objects~\emph{paper-cup} and \emph{plastic-bottle}, our model performs significantly better. This shows our proposed approach has the ability to estimate accurate forces on "never seen" data, as we only collected our dataset on rigid objects. For the \emph{mustard} object, the index fingertip was on the cap, as shown in~Fig.~1 for four of the trials to check the predictions on non-flat contact surfaces. We see that only our method was able to detect placement and successfully release the object.

\section{Conclusion \& Future Work}
\label{sec:conclusion}
We have explored the effect of combining large-scale data from multiple sources with feed-forward neural networks for robustly learning to estimate force from tactile fingertip sensors. We showed significant improvements compared to simple linear models and small-scale neural network methods. We also formulated an optimization scheme to obtain ground truth force from planar pushing of known objects. In future work we will explore the use of these learned force estimates in providing feedback for in-hand manipulation, extending our previous work on in-hand regrasp planning~\cite{sundaralingam2018geometric}. We will additionally extend our planar force estimate formulation to infer dynamic properties of unknown objects through pushing and in-hand manipulation. With the large scale dataset, we hope to also model sensor drift, thereby alleviating the need to tare the signals often in the future.
\section*{Acknowledgment}
We would like to thank Giovanni Sutanto for fabrication of the BioTac mount. We would also like to thank Siddhartha S. Srinivasa, Filipe Veiga and Zhe Su for helpful discussions.

\bibliography{tactile_ref}
\end{document}